\documentclass[lettersize,journal]{IEEEtran}
\usepackage{amsmath,amsfonts}
\usepackage{algorithm}
\usepackage{algorithmicx}
\usepackage{algpseudocode}
\usepackage{multirow,array}
\usepackage[caption=false,font=normalsize,labelfont=sf,textfont=sf]{subfig}
\usepackage{textcomp}
\usepackage{stfloats}
\usepackage{url}
\usepackage{verbatim}
\usepackage{graphicx}
\usepackage{cite}
\usepackage{booktabs}
\usepackage{enumitem}
\usepackage[dvipsnames]{xcolor}
\usepackage[most]{tcolorbox}
\usepackage{float}
\begin{document}

\title{Multi-Task Learning with LLMs for Implicit Sentiment Analysis: Data-level and Task-level Automatic Weight Learning}
\author{Wenna Lai, Haoran Xie, Guandong Xu, Qing Li
\thanks{The research was supported by the Faculty Research Grant (DB24A4) at Lingnan University, Hong Kong. \emph{(Corresponding author: Haoran Xie.)}\\
\indent Wenna Lai is with the Department of Computing, Hong Kong Polytechnic University, Hong Kong SAR (email: winnelai05@gmail.com).\\
\indent Haoran Xie is with the School of Data Science, Lingnan University, Hong Kong (email: hrxie@ln.edu.hk).\\
\indent Guandong Xu is with the School of Computer Science and the Data Science Institute, University of Technology Sydney, Sydney, NSW 2007, and also the Education University of Hong Kong, Hong Kong SAR (e-mail: gdxu@eduhk.hk).\\
\indent Qing Li is with the Department of Computing, Hong Kong Polytechnic University, Hong Kong SAR (e-mail: qing-prof.li@polyu.edu.hk).
}}



\maketitle

\begin{abstract}
Implicit sentiment analysis (ISA) presents significant challenges due to the absence of salient cue words. Previous methods have struggled with insufficient data and limited reasoning capabilities to infer underlying opinions. Integrating multi-task learning (MTL) with large language models (LLMs) offers the potential to enable models of varying sizes to reliably perceive and recognize genuine opinions in ISA. However, existing MTL approaches are constrained by two sources of uncertainty: \textbf{\emph{data-level uncertainty}}, arising from hallucination problems in LLM-generated contextual information, and \textbf{\emph{task-level uncertainty}}, stemming from the varying capacities of models to process contextual information. To handle these uncertainties, we introduce \emph{MT-ISA}, a novel MTL framework that enhances ISA by leveraging the generation and reasoning capabilities of LLMs through automatic MTL. Specifically, \emph{MT-ISA} constructs auxiliary tasks using generative LLMs to supplement sentiment elements and incorporates automatic MTL to fully exploit auxiliary data. We introduce data-level and task-level automatic weight learning (AWL), which dynamically identifies relationships and prioritizes more reliable data and critical tasks, enabling models of varying sizes to adaptively learn fine-grained weights based on their reasoning capabilities. We investigate three strategies for data-level AWL, while also introducing homoscedastic uncertainty for task-level AWL. Extensive experiments reveal that models of varying sizes achieve an optimal balance between primary prediction and auxiliary tasks in \emph{MT-ISA}. This underscores the effectiveness and adaptability of our approach.
\end{abstract}

\begin{IEEEkeywords}
Implicit sentiment analysis, Multi-task learning, Large language models.
\end{IEEEkeywords}

\section{Introduction}

\begin{figure}[h]
\centering
\includegraphics[width=\linewidth]{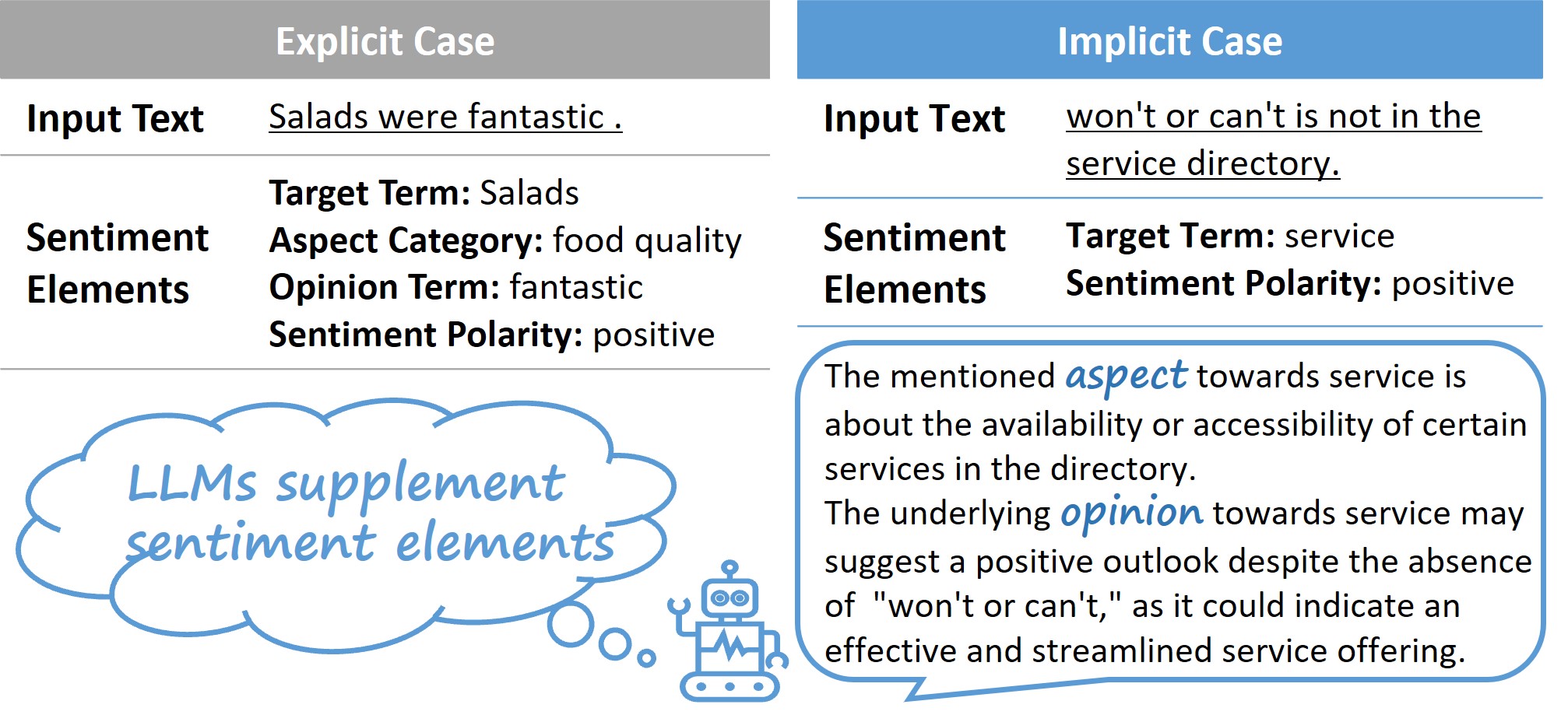}
\caption{The examples illustrate explicit (left) and implicit (right) cases in ABSA. LLMs help supplement sentiment elements in ISA.}
\label{fig:example}
\end{figure}

Aspect-based sentiment analysis (ABSA) aims to identify various sentiment elements, including target terms, aspect categories, opinion terms, and sentiment polarities \cite{Peng_Xu_Bing_Huang_Lu_Si_2020}.  As illustrated in Figure \ref{fig:example}, these sentiment elements are evident in the explicit case to form a complete sentiment picture. Previous dominant research focused on detecting these elements independently. However, identifying a single sentiment element remains insufficient for a comprehensive understanding of aspect-level opinions \cite{Zhang2022ASO}. To delve into more complex real-world scenarios, Li et al. \cite{Li2021LearningIS} shed light on implicit sentiment analysis (ISA) and pointed out that previous studies have paid less attention to ISA, which presents greater challenges due to the absence of salient cue words in expressions. Unlike traditional ABSA tasks, ISA necessitates not only the extraction of dependencies between sentiment elements but also the capture of nuanced sentiment cues conveyed within texts \cite{Zhang2023SentimentAI}.

The key challenges present in ISA are \textbf{\emph{insufficient data}} to support pattern learning for implicit sentiments and \textbf{\emph{limited reasoning capabilities}} of traditional models to incorporate common sense knowledge. In the evolving landscape of SA, researchers have increasingly focused on enhancing techniques for ISA. Li et al. \cite{Li2021LearningIS} pre-trained on a large-scale sentiment-annotated corpus, aiming to align implicit representations with explicit ones. However, their approach demands significant effort in constructing data sources and lacks generalization and interpretability in reasoning deep-level sentiments. Recent advancements in large language models (LLMs) have exhibited their potential in addressing the challenges of ISA, particularly due to their remarkable performance in natural language generation and understanding \cite{Zhang2023SentimentAI, Zhang2024AffectiveCI}. Figure \ref{fig:example} presents an example where LLMs supplement sentiment elements in the implicit case, demonstrating their capabilities for reasoning about sentiments. Fei et al. \cite{Fei_Li_Liu_Bing_Li_Chua} directly conducted Chain-of-Thought (CoT) fine-tuning to extract relevant sentiment elements for discerning implicit orientations in expressions. Although this approach achieved some performance improvements, it necessitates LLMs of a certain scale to exhibit emergent abilities \cite{emergent22} for step-by-step inference, limiting its broader application. Furthermore, LLMs are prone to hallucination problems, which can result in unfaithful reasoning and subsequently impair learning performance\cite{Zhang2023SirensSI, Li2024ASO}. 

Apart from CoT fine-tuning, another prominent technique for integrating LLMs in reasoning is Multi-task learning (MTL), which has shown promise in various NLP tasks by leveraging shared representations across related reasoning tasks to improve overall performance \cite{Collobert2008AUA, zhang-etal-2023-survey}. Notably, MTL with LLMs has demonstrated effectiveness even with smaller models \cite{HoSY23, LI2022ExplanationsFL}. They typically use LLMs for auxiliary data generation and joint learning through MTL. However, traditional MTL approaches often require manual tuning of task weights \cite{HoSY23, LI2022ExplanationsFL}, which can be time-consuming and suboptimal due to two primary sources of uncertainty when applied with LLMs: \textbf{\emph{data-level uncertainty}}, arising from the hallucination problems inherent in LLM-generated contextual information, and \textbf{\emph{task-level uncertainty}}, stemming from the varying capacities of models to digest contextual information. It is critical to enhance the self-knowledge of LLMs \cite{Li2024ASO} and handle these potential uncertainties, enabling models of different sizes to reliably perceive and recognize genuine opinions in ISA.

To tackle the above challenges and effectively harness the potential of LLMs for reliable reasoning, we introduce a novelMTL framework, MT-ISA, which leverages the generation and reasoning capabilities of LLMs,  while also managing inherent uncertainties at both the data and task levels through automatic MTL. In MT-ISA, we construct auxiliary tasks by utilizing generative LLMs to supplement additional sentiment elements. By integrating MTL with LLMs, MT-ISA can fully exploit auxiliary data generated by LLMs, enabling backbone models to capture relationships between sentiment elements and effectively reason about implicit sentiments. To further enhance reasoning learning by managing inherent uncertainties,  we employ data-level and task-level automatic weight learning (AWL) to dynamically adjust the focus on more reliable data and critical tasks, eliminating the need for extensive manual intervention. This allows models of varying sizes to adaptively learn fine-grained weights based on their reasoning capabilities. We explore three strategies for data-level AWL and incorporate homoscedastic uncertainty for task-level AWL, contributing to optimal reasoning learning performance in ISA. 

In general, the key contributions of this work are as follows:
\begin{itemize}
    \item We propose MT-ISA, a novel MTL framework for ISA that leverages the generation and reasoning capabilities of LLMs, while effectively handling inherent data-level and task-level uncertainties through automatic MTL.
    \item MT-ISA incorporates data-level and task-level AWL to dynamically prioritize reliable data and critical tasks, enabling models of varying sizes to adaptively learn fine-grained weights based on their reasoning capabilities without manual intervention.
    \item We investigate three strategies for data-level AWL, including Input (I), Output (O), and Input-Output (I-O) strategies. Extensive experiments demonstrate the efficacy of these strategies in enhancing automatic MTL performance, achieving state-of-the-art results in ISA.
\end{itemize}

The following sections are structured as follows. Section \ref{related_work} will introduce related works in the field of ISA and MTL. Then, Section \ref{methodology} will detail the proposed MT-ISA framework, including auxiliary task construction and automatic weight learning. Subsequently, Section \ref{experiment} will present experimental evaluations conducted on two benchmark datasets. In section \ref{discuss}, we will discuss alternative methods we have compared. Finally, conclusions are drawn in Section \ref{conclusion}.

\section{Related Works}
\label{related_work}
\subsection{Implicit Sentiment Analysis}

Implicit sentiment analysis (ISA) presents greater challenges compared to sentiment classification tasks due to the absence of salient cue words in expressions \cite{Russo_Caselli_Strapparava_2015, Tubishat2018ImplicitAE, Li2021LearningIS, Zong_Xia_Zhang_2021}. Some prior studies addressed ISA by enhancing feature representation at the sentence level\cite{Xu_Wang_Feng_Yang_Zhang, Zhou_Wang_Zhang_He_2021}. In 2021, the research team \cite{Li2021LearningIS} divided the SemEval-2014 Restaurant and Laptop benchmarks into Explicit Sentiment Expression (ESE) slice and Implicit Sentiment Expression (ISE) slice based on the presence of opinion words, to facilitate a more fine-grained aspect-level analysis. Given that Pre-trained Language Models (PLMs) such as BERT \cite{Devlin_Chang_Lee_Toutanova_2019} demonstrate exceptional performance in learning representations for classification tasks, many researchers have utilized PLMs for downstream tasks. They enhanced ISA through data augmentation \cite{wang-etal-2022-contrastive} or advanced deep learning techniques like contrastive learning \cite{Li2021LearningIS}, graph learning \cite{wang-etal-2020-relational}. In addition, casual intervention has proven to be an effective method for exploring implicit relationships \cite{Wang_Zhou_Sun_Ye_Gui_Zhang_Huang_2022}. With the triumph of LLMs in the NLP domain, the research community has experienced a paradigm shift towards the utilization of LLMs \cite{Zhang2023SentimentAI, Zhang2024AffectiveCI}. Ouyang et al. \cite{Ouyang_Yang_Liang_Wang_Wang_Li_2024} employed encoder-decoder style LLMs, with the encoder trained for data augmentation and the decoder for prediction purposes. Fei et al. \cite{Fei_Li_Liu_Bing_Li_Chua} introduced THOR and applied Chain-of-Thought (CoT) fine-tuning in ISA by exploiting the emergent abilities in LLMs. However, THOR requires larger models to fully exhibit in-context learning and reasoning capacities, while smaller models achieve limited performance with CoT reasoning. In comparison, our approach enhances reasoning context using generative LLMs to self-refine \cite{nips/refine} with label intervention for reliability control. Additionally, We capitalize on the strengths of LLMs via MTL employing data-level and task-level AWL to handle inherent uncertainties, which enables backbone models to deliver optimal results regardless of model size.

\subsection{Multi-task Learning}
Multi-task learning (MTL) aims to enhance generalization performance by leveraging shared information across multiple related tasks  \cite{Caruana1997MultitaskL}. It is a promising direction in NLP with various applications including information extraction, natural language understanding, and generation \cite{zhang-etal-2023-survey}. In ABSA, prior studies have explored joint learning of subtasks to improve the performance of sentiment polarity classification \cite{Wang2021MultitaskBF, Schmitt2018JointAA}. Yang et al. \cite{Yang2019AML} developed an MTL model to jointly extract aspects and predict their corresponding sentiments, achieving enhanced performance on both Chinese and English review datasets. Similarly, Yu et al. \cite{10080017} proposed a multi-task learning framework utilizing the pre-trained BERT model as a shared representation layer, but applied more complex tasks for joint learning including aspect-term SA and aspect-category SA tasks. Zhao et al. \cite{ZhaoLCQ23} introduced MTABSA to learn aspect-term extraction and sentiment classification simultaneously, incorporating multi-head attention to associate dependency sequences with aspect extraction. In contrast, our approach focuses on a more granular level of joint task learning, where each sentiment element functions as a subtask and contributes to the primary sentiment polarity prediction task.
In recent years, PLMs like T5 \cite{Raffel2019ExploringTL} have demonstrated exceptional generalization capabilities across numerous NLP tasks through multi-task learning via task prefixes. Models empowered by PLMs have shown promising results \cite{Fei_Li_Liu_Bing_Li_Chua, Ouyang_Yang_Liang_Wang_Wang_Li_2024, Wang2022UnifiedABSAAU, Gou2023MvPMP}. Furthermore, LLMs with larger size have exhibited impressive complex reasoning abilities using Chain-of-Thought (CoT) prompting \cite{Wei_Wang_Schuurmans_Bosma_Chi_Le_Zhou, Kojima_Shixiang_Gu_Reid_Matsuo_Iwasawa, Brown2020LanguageMA, Fu_Peng_Sabharwal_Clark_Khot_2022, Zhang_Zhang_Li_Smola_2022}. Some research has focused on jointly learning reasoning chains with prediction tasks \cite{Rajani_McCann_Xiong_Socher_2019, HoSY23, LI2022ExplanationsFL}. Li et al.\cite{LI2022ExplanationsFL} explored multi-task learning with explanation and prediction, leveraging T5 as the backbone model with instruction learning. However, their approach assigns equal weight to each task, and the auxiliary explanation task may produce unfaithful explanations that could negatively impact prediction tasks. In contrast, our method conducts automatic MTL utilizing data-level and task-level AWL to focus on more reliable data and critical tasks, which achieve optimal performance by learning more fine-grained weights without extensive manual intervention.
\section{Methodology}
\label{methodology}
\begin{figure*}[h!]
\centering
\includegraphics[width=\textwidth]{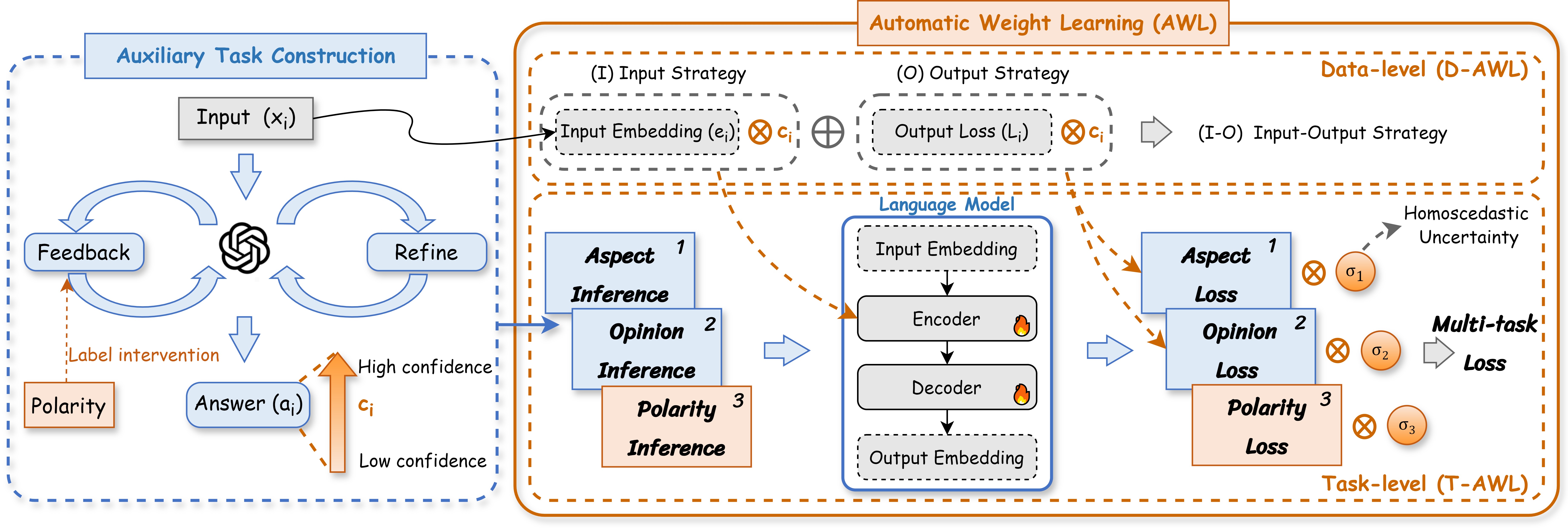}
\caption{The overview of proposed multi-task learning framework MT-ISA. The primary task is polarity inference originating from the given dataset. The auxiliary tasks are constructed by LLM generation using the self-refine strategy with polarity intervention to guide the generation for relevant sentiment elements, including aspect and opinion. The backbone model is trained using multi-task learning with automatic weight learning (AWL), which simultaneously considers auxiliary data confidence for data-level AWL and homoscedastic uncertainty (i.e., task-level uncertainty) for task-level AWL to obtain fine-grained weights and achieve optimal learning performance.}
\label{fig:framework}
\end{figure*}
In this section, we delve into the proposed multi-task learning framework MT-ISA, as illustrated in Figure \ref{fig:framework}. MT-ISA leverages the generation and reasoning capabilities of LLMs through automatic MTL, which dynamically adjusts the weights between primary tasks and auxiliary tasks by concerning the confidence level of data and the importance of tasks. In the beginning, MT-ISA constructs additional sentiment elements as auxiliary tasks to form a comprehensive sentiment picture of a given expression. To prepare relevant sentiment elements conducive to the primary polarity inference task, an off-the-shelf LLM is employed to perform generation using a self-refine strategy with polarity label intervention, ensuring data relevance and quality. Moreover, confidence scores are obtained from the generation process, regarded as indicators of data-level uncertainty. To further enhance learning performance by handling inherent uncertainties that exist when applying MTL with LLMs, we introduce data-level and task-level automatic weight learning (AWL), enabling models of varying sizes to adaptively learn fine-grained weights based on their reasoning
capabilities without manual intervention. The following section will provide a detailed explanation.

\subsection{Task Definition}
Denote a dataset $D = \{(x_i, t_i, y_i)\}^N$, where $1 \le i \le N$, $N$ is the number of data instance. Given an input sentence $x_i$ and its corresponding aspect term $t_i$, where $t_i \subset x_i$, the objective of ISA is to infer the sentiment polarity $y_i$ towards aspect term $t_i$ correctly. The relevant sentiment elements consist of \emph{aspect} $a_i$ and \emph{opinion} $o_i$, which may not be explicitly included in the input sentence $x_i$. In the setting of single-task fine-tuning with prompting approach, the LLM predicts the sentiment polarity $\hat{y_i}$ solely via $\hat{y_i} = argmax \, p(y_i|x_i,t_i)$. This approach may potentially limit the ability of models to capture complete sentiment information in input text without extra information about relevant sentiment elements.
\paragraph{Primary Task}
The objective of ISA is to infer implicit sentiment polarity $y_i$ towards a given target $t_i$ within a given input $x_i$. Therefore, the primary task is polarity inference using a labeled dataset, aiming to predict  $\hat{y_i} = argmax \, p(y_i|x_i,t_i)$. To optimize this task, the cross-entropy loss is calculated between the predicted label $\hat{y_i}$ and the annotated label $y_i$, then we have the prediction loss for sentiment polarity:
\begin{align}
    \mathcal L_{p} = \frac{1}{N} \sum_{i=1}^{N} \ell_{CE}(\hat{y_i}, y_i)
\end{align}


\subsection{Auxiliary Task Construction}
To facilitate the model understanding of implicit sentiment by providing a complete sentiment picture, we leverage generative LLMs to supplement additional sentiment elements that are conducive to enhancing the reasoning capacity for ISA. We construct two subtasks as auxiliary tasks focused on essential sentiment elements including \emph{aspect} and \emph{opinion}. Specifically, we prompt an off-the-shelf LLM using the self-refine strategy with polarity intervention. This approach ensures that the generated sentiment elements are rectified by golden polarity through an iterative refinement process, thereby enhancing the reliability of the generated content. Furthermore, we retrieve the confidence score for generated responses, serving as indicators of data-level uncertainty in Section \ref{d-awl}. The generation process is detailed in Algorithm \ref{refine} and the prompt templates are specified in Figure \ref{fig:case study} with a case study to illustrate the whole process. We use teacher forcing to train our model for auxiliary tasks, minimizing the negative log-likelihood, which is expressed as follows:
\begin{align}
\label{auxilary_loss}
    \mathcal L_{NLL} = - \mathbb{E} log\,p(y|x) = -\mathbb{E} \sum_{t=1}^{T} log\,p(s_t|x,s_{<t})
\end{align}
\begin{align}
   \mathcal L_{a} = \mathcal L_{o} =\mathcal L_{NLL}
\end{align}
where $L_{a}$ and $L_{p}$ are the loss for aspect and opinion auxiliary tasks respectively, $T$ is the length of target sequence $s$, and $s_{<t}$ represents the sequence of outputs before time step $t$. 
\begin{algorithm}
\caption{Auxilary Task Construction}
\label{refine}
\textbf{Given:}  An original dataset $D = \{(x_i, t_i, y_i)\}^N$. An off-the-shelf LLM $\mathcal F$ and max refinement epoch $\mathcal E$. A set of prompt template $\mathcal T$.\\
\textbf{Output:} Sentiment element information including aspect $a_i$ with confidence score $c_{a_i}$ and opinion $o_i$ with confidence score $c_{o_i}$. Auxiliary tasks data $D_a = \{(x_i, t_i, a_i, c_{a_i})\}^N$ and $D_o = \{(x_i, t_i, o_i, c_{o_i})\}^N$.
\begin{algorithmic}[1]
\For{$(x_i, t_i, y_i)$ in $D$} 
\State Set auxiliary tasks data $D_a = \{\}, D_o = \{\}$
\State Set initial prompt $p_i = x_i$
\State Set initial feedback $f_i = None$
\For{$e < \mathcal E$}
\State $p_i = \mathcal T(x_i, t_i, f_i)$ 
\State $a_i, c_{a_i} = \mathcal F(p_i)$ \Comment{infer the aspect}
\State $p_i = \mathcal T(x_i, t_i, a_i, f_i)$
\State $o_i, c_{o_i} = \mathcal F(p_i)$ \Comment{infer the opinion}
\State $p_i = \mathcal T(x_i, t_i, a_i, o_i, f_i)$
\State $\hat{y_i} = \mathcal F(p_i)$ \Comment{infer the polarity}
\If{$\hat{y_i} = y_i$}
\State break
\EndIf
\State $p_i = \mathcal T(x_i, t_i, a_i, o_i, \hat{y_i})$ \Comment{prompt for feedback}
\State $f_i = \mathcal F(p_i)$\Comment{infer the feedback for refinement}
\EndFor
\State $D_a \gets D_a \cup \{(x_i, t_i, a_i, c_{a_i})\}$
\State $D_o \gets D_o \cup \{(x_i, t_i, o_i, c_{o_i})\}$
\EndFor
\end{algorithmic}
\end{algorithm}
\subsection{Data-level Automatic Weight Learning (D-AWL)}
\label{d-awl}
To encourage the model to prioritize data instances with high confidence levels in auxiliary tasks, we adopt three strategies for data-level AWL using the confidence scores obtained from Algorithm \ref{refine}. These strategies help the model better manage uncertainty and noisy data, enabling it to learn more meaningful feature representations by mitigating the negative impact of noisy data on model training and enhancing learning efficiency and effectiveness. We also explored alternative methods for retrieving confidence scores, which are discussed in Section \ref{discuss}. We select the prompt-based method due to its more consistent distribution and superior performance in application. Specifically, we conduct an empirical study on three data-level AWL strategies as follows:

\subsubsection{Input (I) Strategy}
The first strategy is scaling the input embedding according to the data-level confidence scores retrieved from Algorithm \ref{refine}, directly influencing the feature representations fed into the backbone model. Given the model embedding layer $Emb(\cdot)$ and input text $x_i$ with confidence score $c_i$, we adapt the auxiliary task loss in equation (\ref{auxilary_loss}) to get:
\begin{align}
    e_i = c_i \cdot Emb(x_i)  
\end{align}
\begin{align}
\label{input_re}
   \mathcal L_{NLL_{(I)}} = - \frac{1}{N} \sum_{i=1}^{N} \sum_{t=1}^{T} log\,p(s_{i,t}|e_i,s_{i,<t})
\end{align}
where the re-weighting embedding $e_i$ will be fed into the backbone model instead of the original $x_i$ for auxiliary task training.
\subsubsection{Output (O) Strategy }
The second strategy is to re-weight the output loss with data-level confidence scores retrieved from Algorithm \ref{refine} while dealing with data instances equally, which adjusts more attention to data instances with higher confidence levels during optimization. In this way, the auxiliary loss in equation (\ref{auxilary_loss}) becomes:
\begin{align}
\label{output_re}
   \mathcal L_{NLL_{(O)}} = - \frac{1}{N} \sum_{i=1}^{N} \sum_{t=1}^{T} c_i \cdot log\,p(s_{i,t}|x_i,s_{i,<t})
\end{align}
\subsubsection{Input-Output (I-O) Strategy }
The third strategy simultaneously conducts confidence-guided re-weighting towards input embedding and output loss, which combines equation (\ref{input_re}) \& (\ref{output_re}) to get:
\begin{align}
\label{io_re}
   \mathcal L_{NLL_{(I-O)}} = - \frac{1}{N} \sum_{i=1}^{N} \sum_{t=1}^{T} c_i \cdot log\,p(s_{i,t}|e_i,s_{i,<t})
\end{align}
\subsection{Task-level Automatic Weight Learning (T-AWL)}
At the task level, we have each subtask represent one sentiment element. Instead of manually adjusting the weight between tasks, MT-ISA introduces \emph{homoscedastic uncertainty} \cite{KendallG17} as the \emph{task-level uncertainty}\footnote{It is also known as \emph{task-dependent uncertainty} \cite{multi1/KendallGC18} proven to effectively reflect the task inherent uncertainties in the multi-task setting. Notably, this uncertainty remains constant across all input data but varies between tasks.} to capture relative confidence among tasks. Subsequently, we develop an automatic loss function (ALF) for the multi-task loss in MT-ISA and train the model based on $ALF_1$:
\begin{align}
\label{multi1}
   \mathcal L = \frac{1}{{\sigma_1}^2}\mathcal L_{a} + \frac{1}{{\sigma_2}^2}\mathcal L_{o} + \frac{1}{{\sigma_3}^2}\mathcal L_{p} + \sum_{i=1}^{k}log({\sigma_i}^2)
\end{align}

where $\sigma$ is a tunable observation noise parameter that measures \emph{task-level uncertainty}, $log({\sigma}^2)$ is the regularization term, and $k$ is the number of varied tasks. 

However, to avoid negative values during optimization introduced by $log({\sigma}^2)$ when ${\sigma}^2 < 1$, the regularization term can be adapted to $ln({\sigma}^2 + 1)$ as suggested by \cite{multi2/abs-1805-06334}. Then we adapt $ALF_2$ as follows:
\begin{align}
\label{multi2}
   \mathcal L = \frac{1}{{\sigma_1}^2}\mathcal L_{a} + \frac{1}{{\sigma_2}^2}\mathcal L_{o} + \frac{1}{{\sigma_3}^2}\mathcal L_{p} + \sum_{i=1}^{k}ln({\sigma_i}^2 + 1)
\end{align}

In practice, we evaluate and compare both $ALF_1$ and $ ALF_2$ as discussed in section \ref{discuss}. Through the implementation of the above-developed loss function $\mathcal L$ for multi-task fine-tuning, we expect the model to dynamically adjust the task weight and model parameters simultaneously under D-AWL and T-AWL strategies.

\section{Experiment}
\label{experiment}
\subsection{Setup}
\paragraph{\textbf{Datasets and Metrics}}
To evaluate the performance of ISA, We take experiments on Restaurant and Laptop datasets in SemEval-2014 \cite{Pontiki_semeval_2014} and follow the prior work \cite{Li2021LearningIS} splitting annotated data into implicit and explicit classes. The evaluation metrics use accuracy and macro-F1 score. 
\paragraph{\textbf{Baselines}}
We compare our approach with state-of-the-art ABSA baselines and recently reported prompt-based fine-tuning methods. Given that we utilize GPT-4o-mini for auxiliary task construction, we also include its zero-shot performance for reference. Additionally, we compare with MTL baselines to demonstrate the effectiveness of our automatic MTL approach. The baselines are as follows:

\begin{itemize}
\item \textbf{\emph{ABSA Baselines:}}
\begin{itemize}
\item \textbf{BERT+SPC} \cite{Devlin_Chang_Lee_Toutanova_2019}: Directly fine-tunes BERT for sentence pair classification.
\item \textbf{BERT+ADA} \cite{rietzler2019adapt}: Integrates data from other domains for adaptation.
\item \textbf{RGAT} \cite{wang-etal-2020-relational}: Learns both aspect term embedding and dependency relation embedding to obtain more information.
\item \textbf{C$^3$DA} \cite{wang-etal-2022-contrastive}: Creates augmentations by modifying the aspect term and altering the sentiment polarity.
\item \textbf{ISAIV} \cite{Wang_Zhou_Sun_Ye_Gui_Zhang_Huang_2022}: Investigating causal relations for implicit sentiment expressions.
\item \textbf{BERT$_{Asp}$+SCAPT} \cite{Li2021LearningIS}: Uses contrastive learning with external corpora to learn implicit knowledge.
\item \textbf{BERT$_{Asp}$+CEPT} \cite{Li2021LearningIS}: Replaces contrastive learning with cross-entropy loss to post-train BERT with external corpora.
\item \textbf{ABSA-ESA} \cite{Ouyang_Yang_Liang_Wang_Wang_Li_2024}: Augments implicit cases with explicit expressions using encoder-decoder models.
\end{itemize}

\item \textbf{\emph{Prompt-based Fine-tuning:}} 
\begin{itemize}
    \item \textbf{Direct Fine-tune}: Serves as a baseline using standard prompting to question sentiment polarity.
    \item \textbf{InstructABSA} \cite{instructabsa}: Conducts instruction tuning that applicable to all ABSA tasks.
    \item \textbf{THOR} \cite{Fei_Li_Liu_Bing_Li_Chua}: Applies CoT fine-tuning to infer sentiment elements step-by-step.
\end{itemize}

\item \textbf{\emph{MTL Methods:}} 
\begin{itemize}
    \item \textbf{BERT-MTL} \cite{10080017}: Simultaneously learns aspect-term and aspect-category sentiment analysis tasks.
    \item \textbf{MTABSA} \cite{ZhaoLCQ23}: Jointly learns aspect term extraction and aspect polarity classification tasks.
    \item \textbf{MT-Re and MT-Ra} \cite{LI2022ExplanationsFL}: Conduct MTL on LLMs by jointly learning rationales and question-answer pairs of original data. MT-Re derives rationales using reasoning (Re) prompts, whereas MT-Ra employs rationalization (Ra) prompts.
\end{itemize}
\end{itemize}

\paragraph{\textbf{Models and Hyperparameters}}
We use an off-the-shelf model, GPT-4o-mini, for auxiliary task construction applying a self-refine strategy \cite{nips/refine} with polarity label intervention. For MTL, we use Flan-T5\cite{flant5} in encoder-decoder architecture as the backbone model. We test different sizes of Flan-T5 scaling from the base model (250M) to the XXL model (13B) to investigate their performance and behavior under data-level and task-level AWL. As our method applies AWL instead of manual adjustment, the task weights for different subtasks are tunable hyperparameters that are optimized along with model parameters. To maintain more stable training, we adopt $ALF_2$ as the multi-task loss objective to present the main results, and we discuss the results obtained with various $ALF$ in Section \ref{discuss}.

\subsection{Main Results}

\begin{table*}[h] 
  \caption{Main results compared with baselines on Restaurant and Laptop datasets. The results with $^\dagger$ and $^\star$ are obtained from \cite{Li2021LearningIS} and \cite{Ouyang_Yang_Liang_Wang_Wang_Li_2024}, while the other results are self-rerun or self-implemented. In our methods, the subscripts stand for data-level AWL strategies: Input Strategy($I$) and Output Strategy($O$), respectively. The subscripts $A$ and $F$ for evaluation metrics represent the accuracy and macro-F1 score.}
  \label{main_result}
  \centering
  \resizebox{0.85\linewidth}{!}{
  \begin{tabular}{llccc|ccc}
    \toprule[1.2pt]
         \multirow{2}{*}{\hfil Method}& \multirow{2}{*}{\hfil Model} & \multicolumn{3}{c}{Restaurant14} & \multicolumn{3}{c}{Laptop14}\\
     \cmidrule(r){3-8}
     && All$_A$ & All$_F$ & ISA$_A$ & All$_A$ & All$_F$ & ISA$_A$ \\
    \hline
    \bf\emph{- State-of-the-art ABSA baselines}&&&&&&  \\
    BERT-SPC$^\dagger$ \cite{Devlin_Chang_Lee_Toutanova_2019} & BERT-Base (110M)& 83.57 & 77.16 &65.54 & 78.22 & 73.45 & 69.54\\
    BERT-ADA$^\dagger$ \cite{rietzler2019adapt}  & BERT-Base (110M)& 87.14 & 80.05 &65.92 & 78.96 & 74.18 &70.11 \\
    RGAT$^\dagger$ \cite{wang-etal-2020-relational}  & BERT-Base(110M)& 86.60 & 81.35 &67.79 & 78.21 & 74.07 &72.99 \\
    BERT$_{Asp}$ + CEPT$^\dagger$ \cite{Li2021LearningIS}& BERT-Base (110M) &87.50 & 82.07 &67.79 & 81.66 & 78.38 &75.86  \\
    C$^3$DA$^\dagger$ \cite{wang-etal-2022-contrastive}& BERT-Base (110M)&86.93& 81.23& 65.54&80.61& 77.11&73.57\\
    ISAIV \cite{Wang_Zhou_Sun_Ye_Gui_Zhang_Huang_2022}& BERT-Base (110M) & 87.05 & 81.40 &-  & 80.41 & 77.25 &- \\
    BERT$_{Asp}$ + SCAPT$^\dagger$ \cite{Li2021LearningIS}  & BERT-Base (110M)& 89.11 & 83.79 & 72.28& 82.76 & 79.15 & 77.59\\
    ABSA-ESA$^\star$ \cite{Ouyang_Yang_Liang_Wang_Wang_Li_2024} & T5-Base (220M)&88.29&81.74&70.78&82.44&79.34&80.00\\
    RGAT$^\dagger$ \cite{wang-etal-2020-relational}  & BERT-Large (340M) &86.87& 79.99 &68.05 & 80.25 & 76.26 &75.43\\
     
    \hline
    \bf\emph{- Prompt-based fine-tuning}&&&&&& \\
    Direct fine-tune & Flan-T5-Base (250M) & 86.88  & 79.78 &  65.17  & 81.98  & 77.93 &  73.71  \\
    InstructABSA \cite{instructabsa}& Tk-INSTRUCT (200M) & 80.89 & 38.26 & 47.19  & 79.31 & 46.29 &73.71 \\
    THOR \cite{Fei_Li_Liu_Bing_Li_Chua}  & Flan-T5-Base (250M)&87.68&81.10&68.54&81.66& 77.51&74.29\\
     Direct fine-tune &Flan-T5-XXL (11B) &89.29&83.68&75.28 & 81.82 & 77.69 & 75.43\\
    THOR \cite{Fei_Li_Liu_Bing_Li_Chua}  & Flan-T5-XXL (11B)&88.57&82.93&73.03&82.29&78.78&76.57\\ 
    \hline
    \bf\emph{- Zero-shot baselines}&&&&&& \\
    GPT-4o-mini&-&84.82&72.52&61.45&79.94&73.83&64.17\\
    GPT-4o-mini + THOR&-&81.25&62.12&46.07&72.10&55.42&37.38\\
    \hline
    \bf\emph{- Automatic multi-task learning}&&&&&& \\
    
    MT-ISA$_I$ (Ours) & Flan-T5-Base (250M) &88.21&82.45&70.41&82.91&79.86&80.57\\
    MT-ISA$_O$ (Ours)& Flan-T5-Base (250M) &85.89&77.85&66.29&81.98&78.10&75.43\\
    MT-ISA$_I$ (Ours)& Flan-T5-XXL (11B) &91.34&86.71&81.27&85.40&82.73&86.29\\
    MT-ISA$_O$ (Ours)& Flan-T5-XXL (11B) &\bf{92.68}&\bf{88.96}&\bf{84.27}&\bf{85.74}&\bf{83.86}&\bf{88.57}\\
    \bottomrule[1.2pt]
  \end{tabular}
  }
\end{table*}
\subsubsection{Comparison with Baselines}
The main results compared with baseline methods are shown in Table \ref{main_result}. It can be seen that our method, MT-ISA, surpasses the current baselines and achieves state-of-the-art results in ISA, demonstrating the effectiveness of our proposed framework for learning relationships among sentiment elements. Notably, even the base-sized model with the input D-AWL strategy, \emph{MT-ISA$_I$}, outperforms most baselines, with the exception of BERT$_{Asp}$ + SCAPT \cite{Li2021LearningIS}. Although BERT$_{Asp}$ + SCAPT is pre-trained on the large-scale aspect-aware annotated dataset and exhibits strong capabilities in ABSA, our method still shows superior performance on the Laptop dataset. In MT-ISA, we employed GPT-4o-mini for auxiliary task construction. We tested the zero-shot performance of GPT-4o-mini for reference and observed a gap between directly applying a strong model like GPT-4o-mini and fine-tuning a base model like Flan-T5-base in ISA. Furthermore, there is a performance degradation in both overall and implicit results when adopting GPT-4o-mini with the THOR method, where THOR uses CoT prompting to elicit relevant sentiment elements. This suggests that chain-of-thought prompting requires the model to maintain context and coherence over multiple steps, which can be challenging for smaller models with less capacity for direct inference. In contrast, our method infers relevant sentiment elements with polarity inference as auxiliary tasks and applies multi-task learning with D-AWL and T-AWL techniques for more reliable and effective supervision of reasoning learning on primary tasks.

\subsubsection{Comparison with Multi-task Learning methods}
 
To further compare with other multi-task learning methods, we demonstrate the results in Table \ref{multi-task}. We utilize the base size model for fair comparison. Our method \emph{MT-ISA$_I$}  outperforms baseline methods that apply multi-task learning in ABSA. Among these, BERT-MTL \cite{10080017} and MTABSA \cite{ZhaoLCQ23} construct auxiliary tasks from the perspective of different task types conducive to sentiment polarity inference. In contrast, our method builds auxiliary tasks intending to complete the sentiment picture and focus on semantic and sentiment correlation. We also implemented MT-Re and MT-Ra following \cite{LI2022ExplanationsFL}, which jointly learn rationales with prediction tasks and similarly incorporate relevant sentiment element information in the rationales. However, they applied the multi-task loss by simply summing the loss from each task, lacking fine-grained control at both the data and task levels to enhance primary task performance. Overall, our method with a base-sized model significantly surpasses other multi-task learning methods by introducing AWL for data-level and task-level optimization.

\begin{table} 
  \caption{Main results of Accuracy compared with other multi-task learning methods. The results with $^\dagger$ is obtained from \cite{10080017} and \cite{ZhaoLCQ23}.}
  \label{multi-task}
  \centering
  \begin{tabular}{llcc|cc}
    \toprule[1.2pt]
         \multirow{2}{*}{\hfil Method}& \multirow{2}{*}{\hfil Model} & \multicolumn{2}{c}{Restaurant14} & \multicolumn{2}{c}{Laptop14}\\
     \cmidrule(r){3-6}
     && All & ISA & All& ISA \\
    \hline
     BERT-MTL $^\dagger$\cite{10080017}& BERT-Base & 84.60 &-  & 80.30 &- \\
     MTABSA $^\dagger$\cite{ZhaoLCQ23}& BERT-Base & 86.88 &-  & 80.56 &- \\
    MT-Re \cite{LI2022ExplanationsFL}& Flan-T5-Base & 86.25 & 63.67 & 81.35 &77.14\\
    MT-Ra \cite{LI2022ExplanationsFL}& Flan-T5-Base & 86.96 & 64.05 & 82.29&76.00\\
   
    \hline
     MT-ISA$_I$ & Flan-T5-Base &\bf{88.21}&\bf{70.41}&\bf{82.91}&\bf{80.57}\\
    \bottomrule[1.2pt]
  \end{tabular}
\end{table}

\subsubsection{Comparison with D-AWL Strategies}
\begin{figure*}[h!t]
\centering
\includegraphics[width=\textwidth]{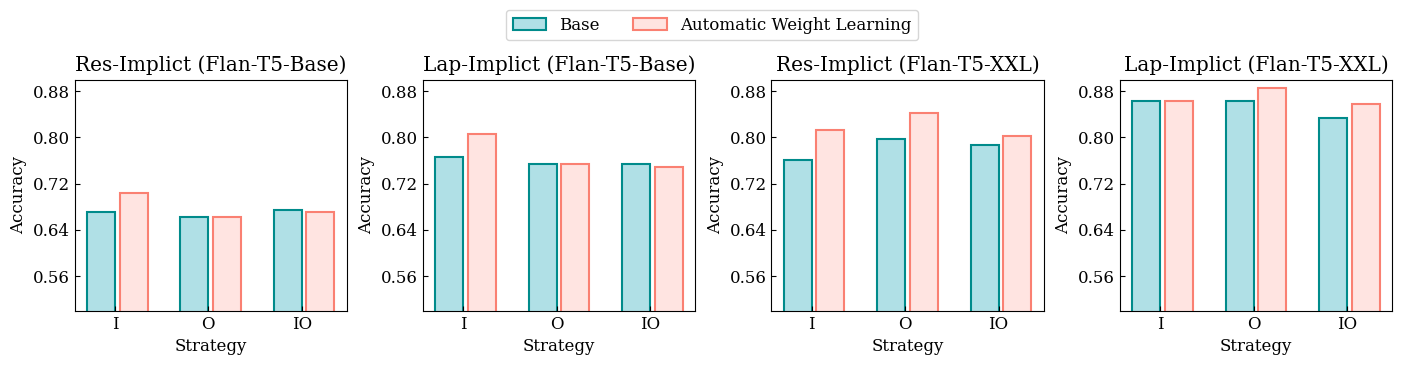}
\caption{Comparision of performance with different D-AWL strategies, including input (I), output (O), and input-output (I-O) strategies. The metric uses the accuracy of implicit datasets.}
\label{fig:reweight}
\end{figure*}

We evaluate and compare different D-AWL strategies from two perspectives, one is \emph{learned task weights} as shown in Table \ref{task weight}, and another is the \emph{accuracy performance} in ISA as illustrated in Figure \ref{fig:reweight}. Results for both base-size and XXL-size models are included. It is observed that the base-size model learns more effectively with the input data-level AWL strategy while the XXL-size model performs better with the output data-level AWL strategy across both datasets. This is because the base-size models with less reasoning capacity may not be capable of digesting the auxiliary information as effectively as XXL-size models, leading them to assign less weight to auxiliary tasks. They benefit more from input-level adjustments which help the model focus on the most relevant features, compensating for its limited capacity to process and prioritize information. While XXL-size models with greater capacity can handle complex input data more effectively, all the strategies assign similarly higher weights for auxiliary tasks compared to that applied in base-size models. Output data-level AWL strategy allows XXL-size models to fine-tune their predictions by emphasizing more confident outputs, leveraging their ability to process and learn from a broader range of data. When it comes to the input-output strategy, it generally lags behind the other two strategies regardless of model size. It is possibly due to over-adjustment, which introduces excessive attention or neglect of certain data by combining both input strategies. In general, these data-level AWL strategies can be conducive to preventing over-fitting problems during training and our method with fine-grained automatic weight learning enables models of different sizes to select the most suitable strategy and achieve optimal performance

\begin{table} 
  \caption{Tasks Weight Learned using Different Data-level Automatic Weight Learning (D-AWL) Strategies in MT-ISA.}
  \label{task weight}
  \normalsize
  \centering
  \resizebox{\linewidth}{!}{
  \begin{tabular}{lccc|ccc}
    \toprule[1.2pt]
         \multirow{2}{*}{\hfil Strategy}& \multicolumn{3}{c}{Restaurant14} & \multicolumn{3}{c}{Laptop14}\\
     \cmidrule(r){2-7}
     &Polarity & Aspect & Opinion & Polarity & Aspect & Opinion \\
    \hline
    Base &0.4&0.3&0.3&0.4&0.3&0.3\\
    \hline
    \multicolumn{3}{l}{\bf\emph{· Flan-T5-Base}}&&&& \\
    D-AWL$_I$  &\bf{0.73}&\bf{0.14}&\bf{0.14}&\bf{0.54}&\bf{0.23}&\bf{0.23}\\
    D-AWL$_O$ & 0.43 & 0.28 & 0.28 & 0.41&0.29&0.29\\
    D-AWL$_{I-O}$& 0.73 & 0.14 & 0.14  & 0.41&0.29&0.29\\
    \hline
    \multicolumn{3}{l}{\bf\emph{· Flan-T5-XXL}} &&&& \\
    D-AWL$_I$  &0.34&0.33&0.33&0.34&0.33&0.33\\
    D-AWL$_O$ & \bf{0.36}&\bf{0.32}&\bf{0.32} &\bf{0.39}&\bf{0.31}&\bf{0.31}\\
    D-AWL$_{I-O}$& 0.43&0.28&0.28  &0.39&0.31&0.31\\
    \bottomrule[1.2pt]
  \end{tabular}
  }
\end{table}

\subsubsection{Model Size Effect}
\begin{figure*}[h!t]
\centering
\includegraphics[width=\textwidth]{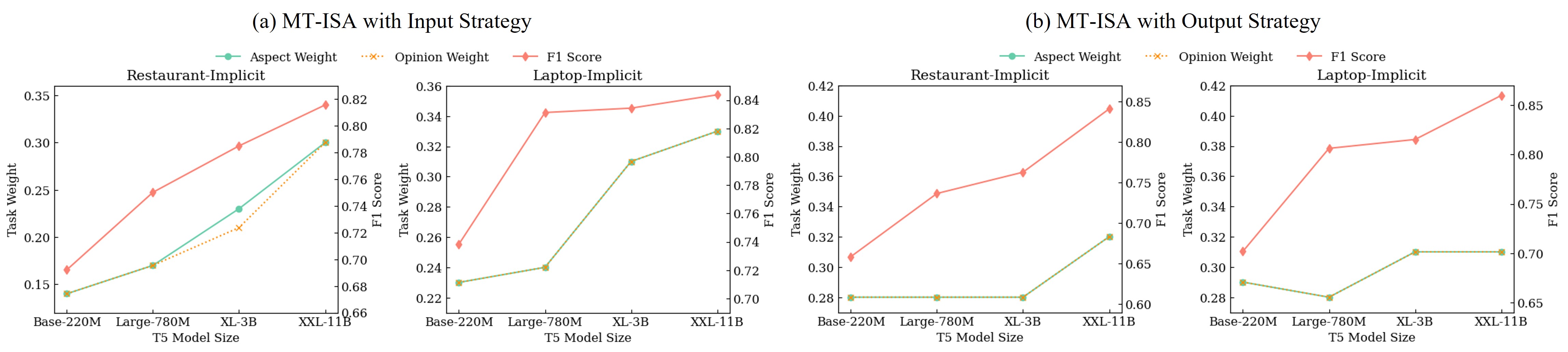}
\caption{The model size effect for MT-ISA with input and output D-AWL strategies. Both auxiliary task weights and implicit F1 score are compared.}
\label{fig:model size}
\end{figure*}

Figure \ref{fig:model size} illustrates the effect of model size on MT-ISA with input and output D-AWL strategies. Both auxiliary task weights and implicit F1 score are compared. In most cases, the assigned weights for aspect and opinion auxiliary tasks overlap because they share the same construction process and task type. As model size increases, the performance of MT-ISA improves regardless of the data-level AWL strategy employed. This is because larger models possess greater reasoning capacity, allowing them to benefit more from auxiliary tasks. For \emph{MT-ISA$_I$}, a similar trend is observed with the assigned task weights for auxiliary tasks. While \emph{MT-ISA$_O$} shows slight changes in task weights as model size increases. Since the ability to handle different complex input data can improve as model size increases under input D-AWL strategy, while output D-AWL strategy maintains a focus on refining predictions, the relative importance of auxiliary tasks remains stable. 

\subsection{Ablation Study}
We conduct an ablation study for the proposed MT-ISA framework to investigate the effects of D-AWL and T-AWL. The best-performing models were evaluated: MT-ISA$_I$ with the base-size model, and MT-ISA$_I$ with the XXL-size model. As shown in Table \ref{ablation}, MT-ISA without D-AWL exhibits a performance degradation of nearly $2\%$, with an even greater drop observed in the Restaurant ISA using the XXL-size model. This underscores the effectiveness of the D-AWL strategy in adjusting data-level attention and enhancing effective learning during multi-task fine-tuning. However, the performance decline of MT-ISA without T-AWL is more pronounced than without D-AWL. Given that multi-task learning can be sensitive to task weights, MT-ISA relying on grid search for optimal task weights may fail to achieve the fine-grained task weight optimization provided by T-AWL. T-AWL plays a crucial role in MT-ISA by balancing different tasks and adapting effectively to models of various sizes. Furthermore, when MT-ISA is applied without the assistance of D-AWL and T-AWL, where we refer collectively to AWL in the table, the performance decreases dramatically with a maximum drop of nearly $9$ points in Restaurant ISA. This highlights the mutually reinforcing relationship between D-AWL and T-AWL, where D-AWL supplements data-level adjustment and T-AWL manages task-level control accounting for the data-level effects of D-AWL.

\begin{table} 
  \caption{Ablation Study for MT-ISA with the macro-F1 score metric. The abbreviations represent Data-level Automatic Weight Learning (D-AWL) and Task-level Automatic Weight Learning (T-AWL).}
  \label{ablation}
  \centering
  \resizebox{\linewidth}{!}{
  \begin{tabular}{llcc|cc}
    \toprule[1.2pt]
         \multirow{2}{*}{\hfil Method}& \multirow{2}{*}{\hfil Model} & \multicolumn{2}{c}{Restaurant14} & \multicolumn{2}{c}{Laptop14}\\
     \cmidrule(r){3-6}
     && All & ISA & All& ISA \\
    \hline
    MT-ISA$_I$ & \multirow{4}{*}{\hfil Flan-T5-Base} &\bf{82.45}&\bf{69.21}&\bf{79.86}&\bf{73.83}\\
    - w/o D-AWL &  & 80.60 & 67.04 & 77.82& 71.38\\
    - w/o T-AWL&  & 80.28 & 66.91  & 77.80 & 70.56\\
    - w/o L&  & 74.89 & 60.43  & 77.76 & 69.98\\
    \hline
     MT-ISA$_O$ & \multirow{4}{*}{\hfil Flan-T5-XXL} &\bf{88.96}&\bf{84.11}&\bf{83.86}&\bf{85.99}\\
    - w/o D-AWL &  & 87.73& 80.23 & 82.19&85.08\\
    - w/o T-AWL&  & 85.78 &79.92  & 82.91 &82.76\\
    - w/o AWL&  & 84.71 &77.36  & 82.23 &82.10\\
    \bottomrule[1.2pt]
  \end{tabular}
  }
\end{table}

\subsection{Case Study}
\begin{figure*}[h]
\centering
\includegraphics[width=\textwidth]{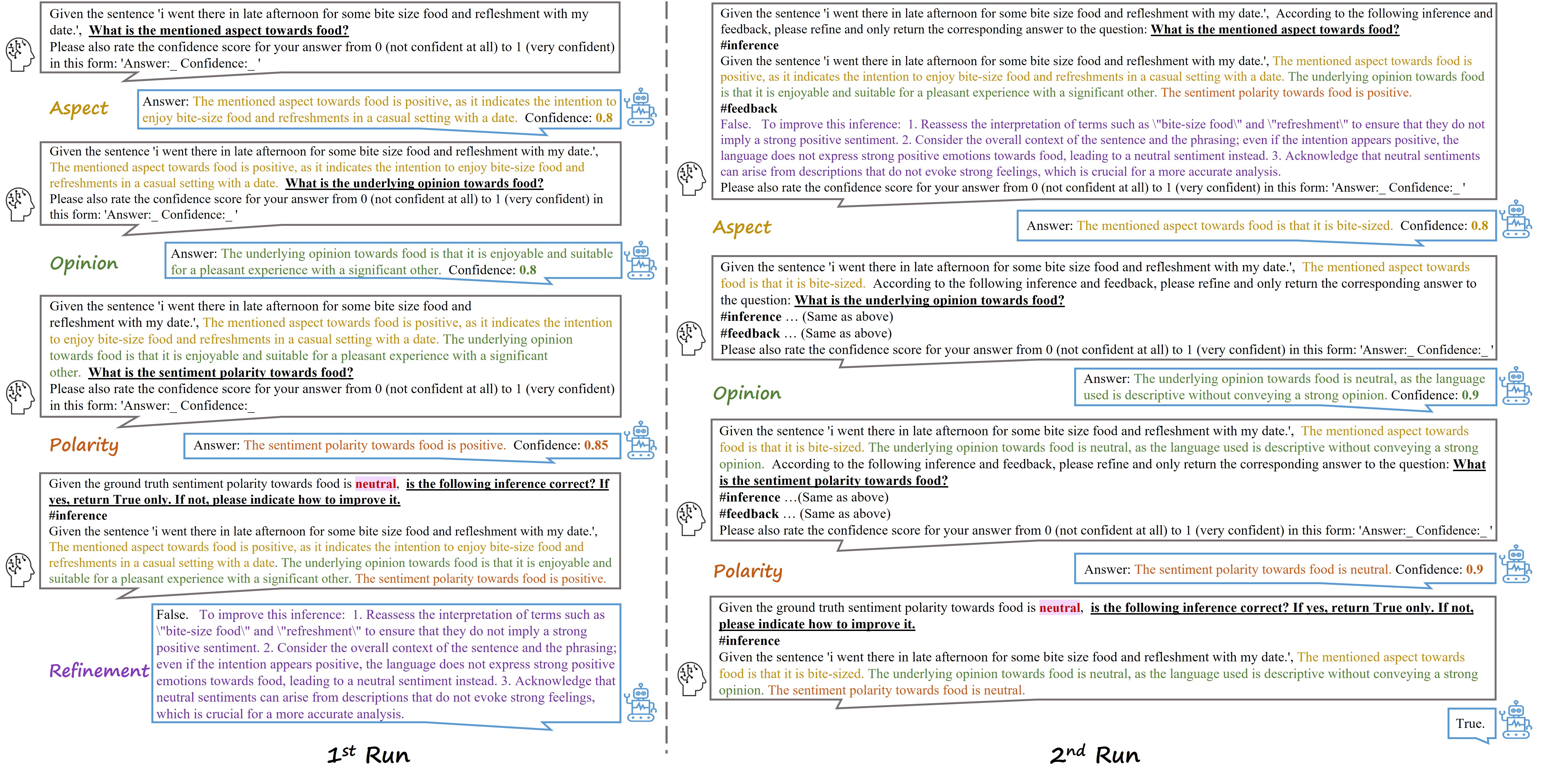}
\caption{An example illustrates Algorithm \ref{refine} designed for constructing auxiliary data. This process generates both aspect $a$ and opinion $o$ using a self-refine strategy with gold label intervention. This example reaches a consensus with GPT-4o-mini after two runs.}
\label{fig:case study}
\end{figure*}

This section presents a case study for auxiliary task construction aimed at restoring aspect $a$ and opinion $o$ in ISA, as illustrated in Figure \ref{fig:case study}. This example involves two rounds of dialogue to achieve consensus with an off-the-shelf LLM, GPT-4o-mini. During the process of inferring sentiment elements, the generated auxiliary data is appended to subsequent questions to serve as a reference. Additionally, the confidence score is elicited with each response. Through the implementation of a self-refine strategy with gold label intervention, GPT-4o-mini effectively refines its responses to align with the ground truth sentiments. Furthermore, the confidence scores exhibit a slight increase following this refinement process. This observation is consistent with the distribution of confidence scores discussed in Section \ref{discuss}.

\section{Discussion}
\label{discuss}
We propose a multi-task learning framework MT-ISA, leveraging generative LLMs for auxiliary data construction to restore the complete sentiment picture. During optimization, we conduct D-AWL and T-AWL to enhance reliable reasoning and achieve optimal performance in ISA. Given the critical roles of data confidence score in D-AWL and automatic loss function in T-AWL, this section will discuss confidence score distribution in part \ref{score} and automatic loss function in part \ref{alf}. We will evaluate alternative approaches and provide an in-depth analysis of their impact on the overall performance of the framework.

\subsection{Confidence Score Distribution}
\label{score}
In the MT-ISA framework, we employ a prompt-based approach to derive confidence scores for data instances. The dominant research elicits confidence scores with prompt-based \cite{li-etal-2024-think} or training-based approach.  Figures \ref{fig:confidence} (a) and (b) illustrate the distribution of confidence scores across two benchmark datasets. Both datasets exhibit a similar distribution, with confidence scores ranging from $0.5$ to $1.0$, and the majority falling between $0.8$ and $1.0$. This trend is attributed to the tendency of the model to assign higher scores to its answer-verified results following several epochs of self-refinement with label intervention. To compare with alternative confidence estimation methods, we investigate two additional approaches using the Laptop dataset, as depicted in Figures \ref{fig:confidence} (c) and (d):
\begin{figure*}[h!t]
\centering
\includegraphics[width=\textwidth]{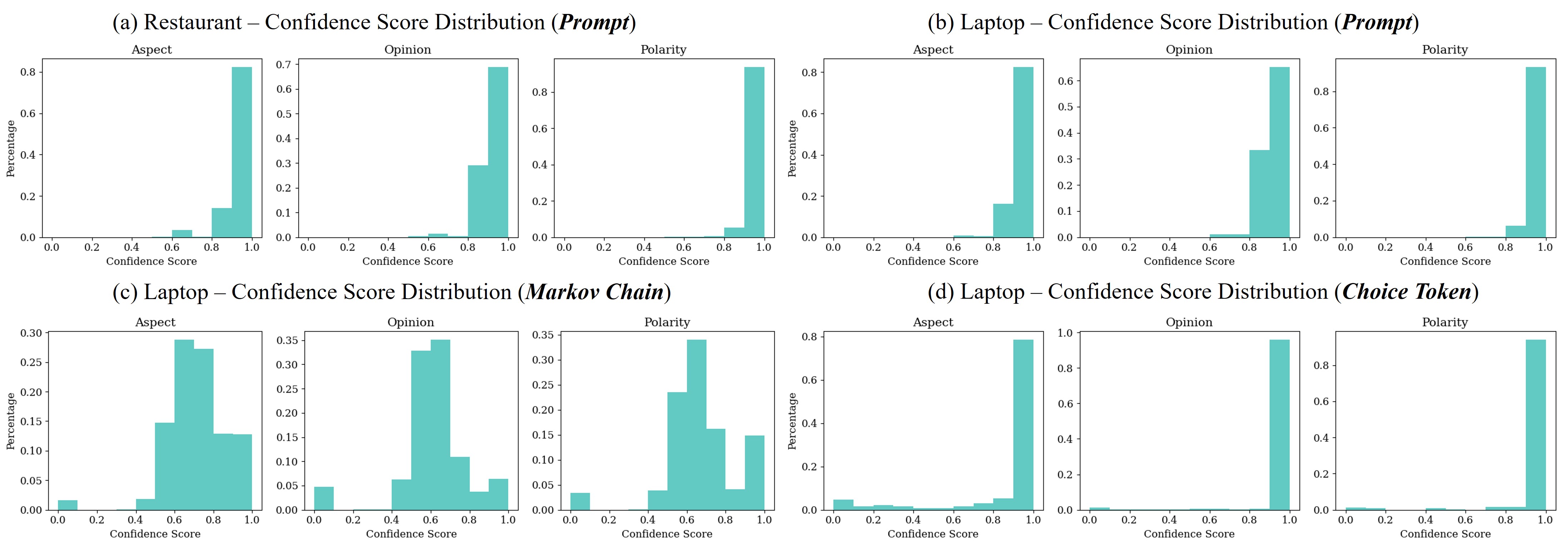}
\caption{The distribution of confidence score retrieved with different methods:(a) Restaurant dataset with Prompt-based method, (b) Laptop dataset with Prompt-based method, (c) Laptop dataset with Markov Chain method, (d) Laptop dataset with Choice Token method.}
\label{fig:confidence}
\end{figure*}
\begin{itemize}
\item \emph{Markov Chain:} This method treats generated sequences as a Markov Chain decision process, calculating the probability of each sequence by summing token-level probabilities and normalizing them to obtain a sequence-level confidence score.
\item \emph{Choice Token:} This method prompts the model to solve a binary classification problem after each generation. For example, the model is prompted to answer: "Is the provided answer reasonable? \#(A) reasonable (B) unreasonable", then the token-level probability for the selected choice generated by the model is used as the confidence score.
\end{itemize}
From the confidence score distribution, the \emph{Markov Chain} approach predominantly yields scores in the range of $0.5$ to $1.0$. In contrast, the \emph{Choice Token} method produces scores mostly between $0.8$ and $1.0$, aligning closely with the prompt-based method depicted in Figure \ref{fig:confidence} (b). However, both the \emph{Markov Chain} and \emph{Choice Token} distributions exhibit a long-tail effect, where low confidence scores can lead to poor generation outcomes when input strategy is applied. Since the backbone models are pre-trained with a stable input distribution, very low confidence scores for input embedding re-weighting may result in the loss of significant information. This distortion can cause the model to misinterpret the input, leading to outputs that are nonsensical or garbled. In practice, we clip the scores to a minimum of $0.5$, assuming that this threshold sufficiently reflects uncertainty to a certain extent. The performance results are presented in Table \ref{diff-confidence}. It is evident that the prompt-based method, with its more consistent confidence estimation, demonstrates superior performance compared to the other two methods. Even when identical auxiliary task weights are learned, variations in prediction performance persist. This observation underscores the critical importance of confidence calibration in facilitating data-level AWL in MT-ISA. Moreover, it highlights the effectiveness of the prompt-based method in eliciting reliable confidence scores during the self-refinement process.

\begin{table} 
  \caption{Comparison of performance on Laptops14 using different methods for retrieving confidence score. The metric uses the macro-F1 score. \textbf{AW} represents auxiliary task weight for aspect/opinion task.}
  \label{diff-confidence}
  \centering
  \begin{tabular}{lllccc}
    \toprule[1.2pt]
         \multirow{2}{*}{\hfil Method}&\multirow{2}{*}{\hfil Confidence}& \multirow{2}{*}{\hfil Model }  & \multicolumn{3}{c}{Laptop14}\\
     \cmidrule(r){4-6}
    & && All& ISA&AW \\
    \hline
    \multirow{3}{*}{\hfil MT-ISA$_I$} & Prompt & \multirow{3}{*}{\hfil Flan-T5-Base}  & \bf{79.86}&\bf{73.83} & \bf{0.23}\\
    &Markov Chain&   & 77.19&70.89&0.23\\
    &Choice Token &  & 76.60 &71.48 & 0.30\\
    \hline
    \multirow{3}{*}{\hfil MT-ISA$_O$} & Prompt & \multirow{3}{*}{\hfil Flan-T5-XXL}  & \bf{83.86}&\bf{85.99}& \bf{0.31}\\
    &Markov Chain&  & 82.81&84.49&0.31\\
    &Choice Token &  & 83.24 &85.47 &0.31\\
    \bottomrule[1.2pt]
  \end{tabular}
\end{table}

\subsection{Automatic Loss Function}
\label{alf}
\begin{table}
  \caption{Comparison of performance when applying T-AWL with $ALF_1$ and $ALF_2$ in MT-ISA. The metric uses the macro-F1 score. \textbf{AW} represents auxiliary task weight for aspect/opinion task.}
  \label{autoloss}
  \centering
  \begin{tabular}{llccc|ccc}
    \toprule[1.2pt]
         \multirow{2}{*}{\hfil Method}&\multirow{2}{*}{\hfil Loss}& \multicolumn{3}{c}{Restaurant14}& \multicolumn{3}{c}{Laptop14}\\
     \cmidrule(r){3-8}
    & & All& ISA& AW & All& ISA & AW\\
    \hline
    \multicolumn{3}{l}{\bf\emph{· Flan-T5-Base}}&&&& \\
    \multirow{2}{*}{\hfil  MT-ISA$_I$} & $ALF_1$  & 81.81 & 68.34 & 0.08 & 79.41&\textbf{74.52} & 0.21\\
    &$ALF_2$ &  \textbf{82.45} & \textbf{69.21}& 0.14  & \textbf{79.86}&73.83& 0.23\\
    \hline
    \multicolumn{3}{l}{\bf\emph{· Flan-T5-XXL}}&&&& \\
     \multirow{2}{*}{\hfil MT-ISA$_O$} & $ALF_1$   &85.88& 79.32 & 0.32 &\textbf{84.81}&83.85 & 0.32\\
    &$ALF_2$  & \textbf{88.96}&\textbf{84.11} & 0.32 & 83.86&\textbf{85.99} &0.31\\
    \bottomrule[1.2pt]
  \end{tabular}
\end{table}
To facilitate T-AWL in multi-task scenarios, we extend the automatic loss function proposed by \cite{multi1/KendallGC18} and \cite{multi2/abs-1805-06334} to our multi-task loss. The primary distinction between these approaches lies in the regularization term, as shown in equations \ref{multi1} and \ref{multi2}. However, their impact during the training process has not been previously explored. In this section, we compare $ALF_1$ with $ALF_2$ for T-AWL and evaluate their performance in the ISA scenario.  The results are presented in Table \ref{autoloss}. It is evident that MT-ISA with $ALF_2$ generally outperforms MT-ISA with $ALF_1$ across most scenarios. It is because $ALF_1$ carries the risk of introducing negative values when $\sigma$ is less than $1$, potentially leading to instability in the training process or undesirable behavior. As observed in the results of MT-ISA$_I$ with $ALF_1$ in the Restaurant dataset, the learned weights for auxiliary tasks tend to approach very small values. This may be due to aggressive adjustments without enforcing a positive value in the regularization term during training, causing the model to converge to a trivial solution. Even when both loss functions learn the same weights for auxiliary tasks, MT-ISA with $ALF_1$ demonstrates inferior performance. This is because the presence of negative values in the loss function can lead to erratic gradients or oscillations during training, affecting convergence behavior and resulting in suboptimal local minima. Overall, MT-ISA is sensitive to task weights and $ALF_2$ provides more stable training results, achieving optimal performance by effectively leveraging auxiliary tasks.

\section{Conclusion}
\label{conclusion}
In conclusion, we introduce a novel MTL framework, MT-ISA, designed to reason genuine underlying opinions in ISA by leveraging the generation and reasoning abilities of LLMs with automatic MTL. MT-ISA amically adjusts weights in MTL according to data-level and task-level uncertainties, enabling models of varying sizes to learn fine-grained weights based on their reasoning capabilities adaptively. We utilize an off-the-shelf LLM for auxiliary task construction, incorporating a self-refinement strategy with polarity label intervention to enhance the reliability of sentiment reasoning, where confidence scores are derived for responses to reflect data-level uncertainty. Our exploration includes threegies for data-level AWL, which are integrated with homoscedastic uncertainty for task-level AWL. MT-ISA demonstrates state-of-the-art performance in ISA, highlighting its efficacy. Additionally, it is noteworthy that models of varying sizes exhibit distinct preferences and influences, yet they achieve optimal performance when employing the dynamic learning strategy. This finding underscores the robustness and adaptability of MT-ISA, suggesting its potential for broader applications.


\bibliography{IEEEabrv, main}

\vspace{100pt}

\begin{IEEEbiography}[{\includegraphics[width=1in, height=1.25in,clip,keepaspectratio]{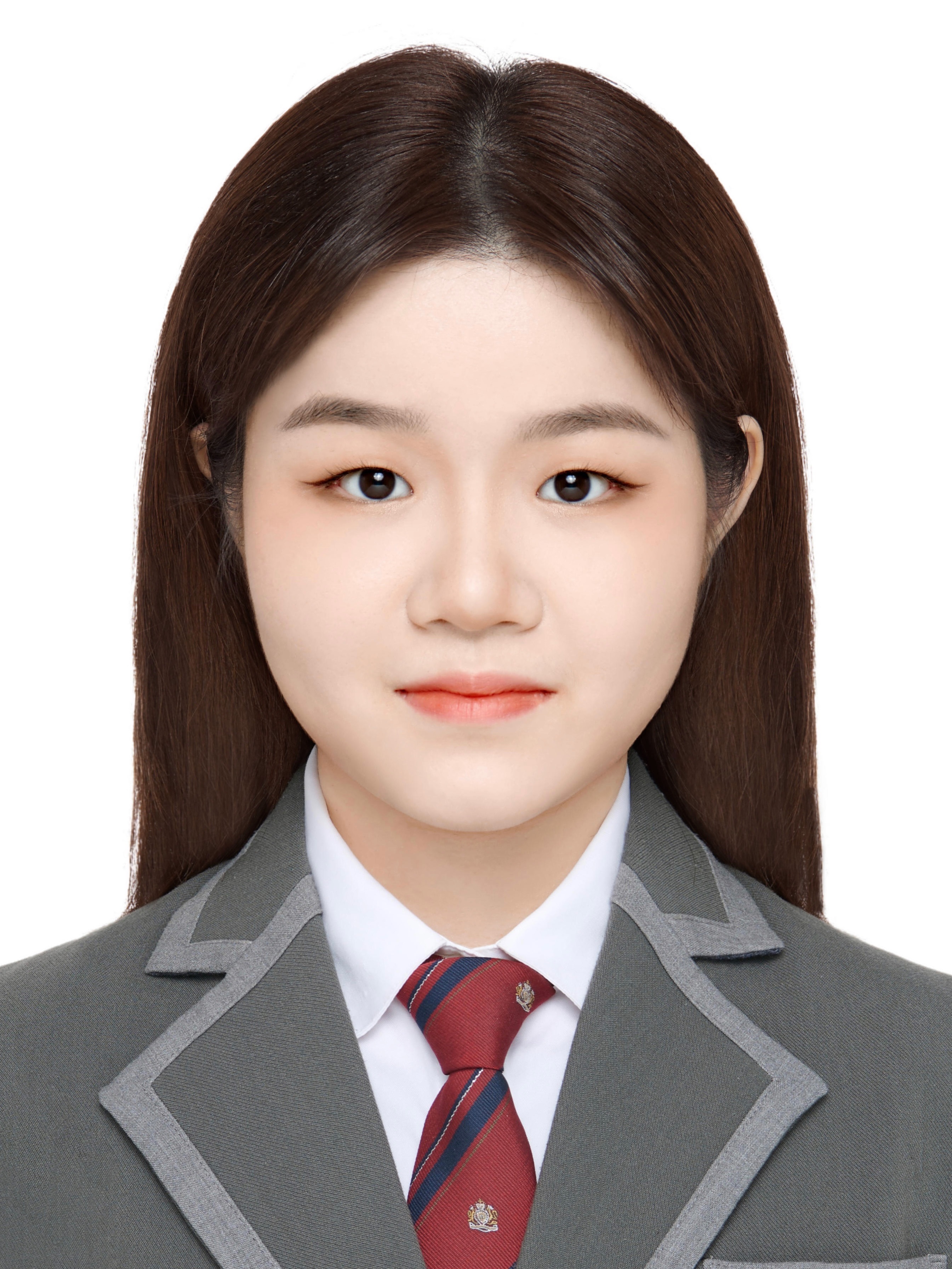}}]{Wenna Lai}
is currently a Ph.D. student at the Department of Computing, Hong Kong Polytechnic University, under the supervision of Prof. Qing Li. She has been working closely with Prof. Guandong Xu at the School of Computer Science, University of Technology Sydney, and Prof. Haoran Xie at the School of Data Science, Lingnan University, Hong Kong. Before that, she received her Master's degree in the Department of Electrical and Computer Engineering from the National University of Singapore. Her research interests include Affective Computing and NLP for Social Good.
\end{IEEEbiography}
\vspace{-1em}
\begin{IEEEbiography}[{\includegraphics[width=1in,height=1.25in,clip,keepaspectratio]{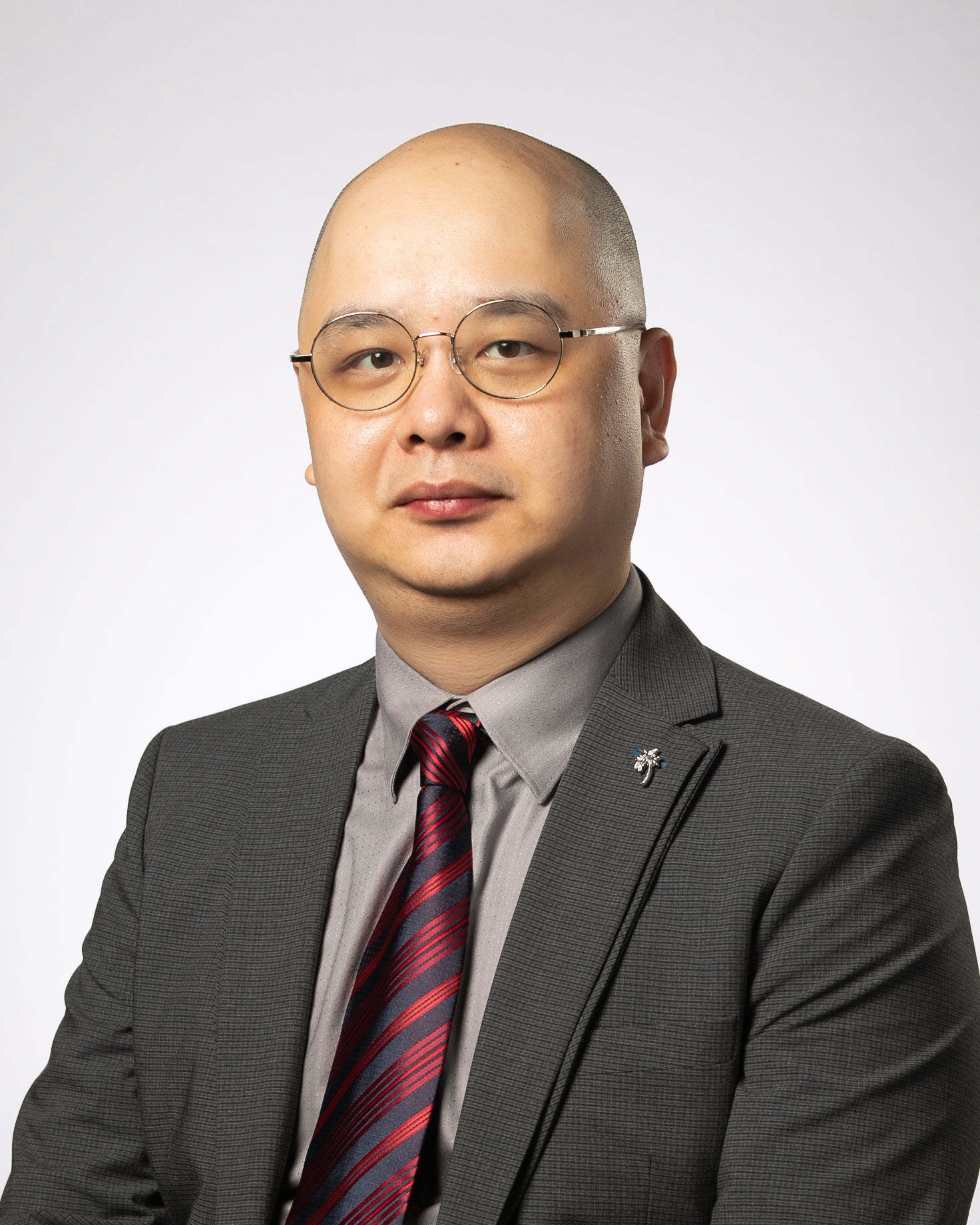}}]{Haoran Xie} (Senior Member, IEEE)
received a Ph.D. degree in Computer Science from City University of Hong Kong and an Ed.D degree in Digital Learning from the University of Bristol. He is currently a Professor and the Person-in-Charge at the Division of Artificial Intelligence, Director of LEO Dr David P. Chan Institute of Data Science, and Acting Associate Dean of the School of Data Science, Lingnan University, Hong Kong. His research interests include natural language processing, large language models, language learning, and AI in education. He has published 423 research publications, including 247 journal articles. He is the Editor-in-Chief of Natural Language Processing Journal, Computers \& Education: Artificial Intelligence, and Computers \& Education: X Reality. He has been selected as the World's Top 2\% Scientists by Stanford University.
\end{IEEEbiography}
\vspace{-1em}
\begin{IEEEbiography}[{\includegraphics[width=1in,height=1.25in,clip,keepaspectratio]{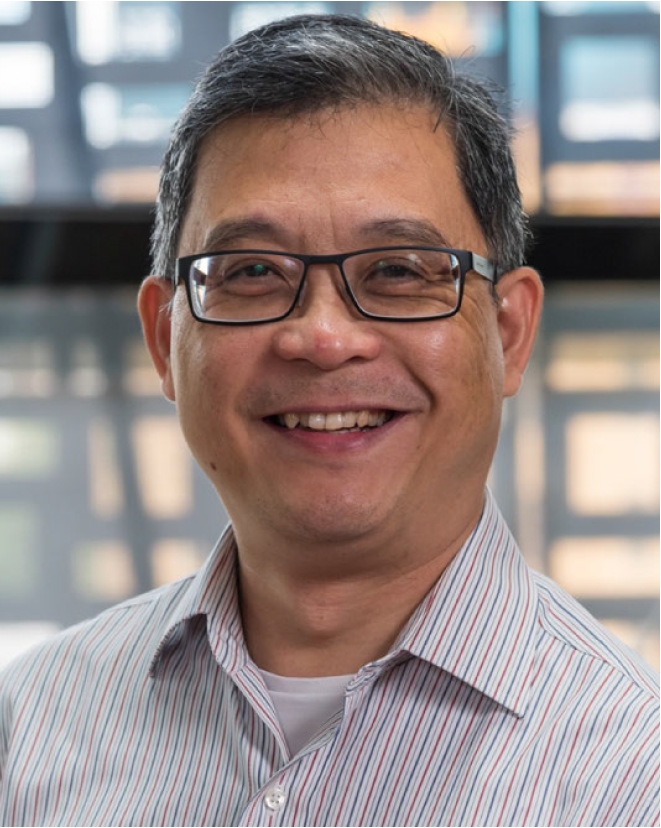}}]{Guandong Xu} (Member, IEEE) received the Ph.D. degree in computer science from Victoria University, Melbourne, VIC, Australia, in 2009. He is currently a Professor and a Program Leader at the School of Computer Science and Data Science Institute, University of Technology Sydney, Sydney, NSW, Australia. His research interests include data science, data analytics, recommender systems, web mining, user modeling, NLP, social network analysis, and social media mining.
\end{IEEEbiography}
\vspace{-1em}
\begin{IEEEbiography}[{\includegraphics[width=1in,height=1.25in,clip,keepaspectratio]{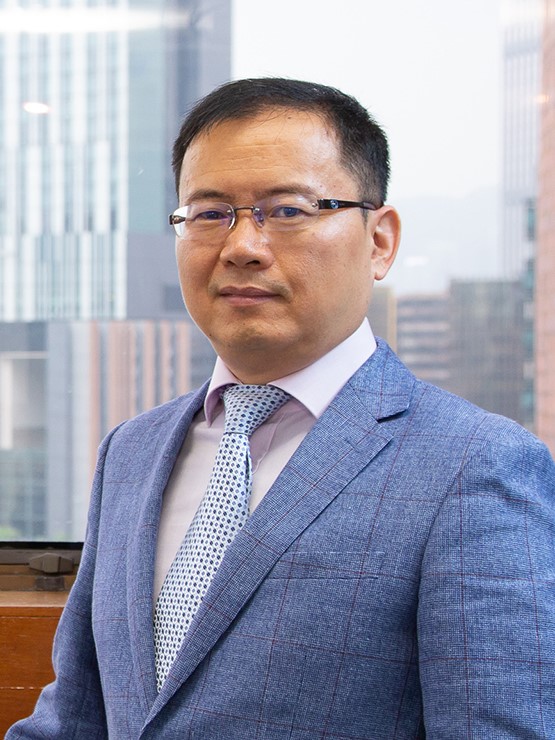}}]{Qing Li} (Fellow, IEEE) received the B.Eng. degree in Computer Science from Hunan Univeristy, Hunan, China, in 1982, and the M.S. and Ph.D. degrees in Computer Science from the University of Southern California, LA, California, USA, in 1985 and 1988, respectively. Qing Li is a Chair Professor and Head at the Department of Computing, The Hong Kong Polytechnic University. His research focuses on data science, web mining, and artificial intelligence. He is a Fellow of IET, a Fellow of IEEE, a member of ACM SIGMOD and IEEE Technical Committee on Data Engineering. He is the chairperson of the Hong Kong Web Society, and is a steering committee member of DASFAA, ICWL, and WISE Society.
\end{IEEEbiography}

\vfill

\end{document}